\documentclass{article}

\usepackage{arxiv}

\usepackage[utf8]{inputenc} 
\usepackage[T1]{fontenc}    
\usepackage{url}            
\usepackage{booktabs}       
\usepackage{amsfonts}       
\usepackage{nicefrac}       
\usepackage{microtype}      
\usepackage{graphicx}
\usepackage{natbib}
\setcitestyle{authoryear,open={(},close={)}} 
\usepackage{hyperref}
\usepackage{doi}
\usepackage{booktabs}
\usepackage{multirow}
\usepackage{makecell} 
\usepackage{xcolor}
\usepackage{adjustbox}

\title{Implementing engrams from a machine learning perspective: the relevance of a latent space}

\date{July 24, 2024}	
\author{
    \href{https://orcid.org/0000-0001-7914-8494}{\includegraphics[scale=0.06]{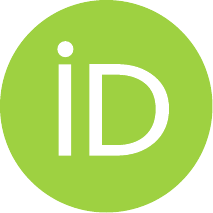}\hspace{1mm}Jesus Marco de Lucas} \\
   \\
    \texttt{jesus.marco@csic.es}
    \\
    Advanced Computing and e-Science Group\\
    Instituto de Física de Cantabria (IFCA) CSIC-Universidad de Cantabria, Santander, ES 39005, SPAIN\\
}

\renewcommand{\headeright}{Short Comment}
\renewcommand{\undertitle}{Short Comment}
\renewcommand{\shorttitle}{\textit{arXiv} Implementing Engrams: Latent Space}

\hypersetup{
pdftitle={Implementing Engrams: Latent Space},
pdfsubject={q-bio.NC, q-bio.QM},
pdfauthor={Jesus.~Marco},
pdfkeywords={Engrams, Autoencoders, Prediction by Matching},
}

\begin{document}
\maketitle

\begin{abstract}

In our previous work, we proposed that engrams in the brain could be biologically implemented as autoencoders over recurrent neural networks. These autoencoders would comprise basic excitatory/inhibitory motifs, with credit assignment deriving from a simple homeostatic criterion. This brief note examines the relevance of the latent space in these autoencoders. We consider the relationship between the dimensionality of these autoencoders and the complexity of the information being encoded. We discuss how observed differences between species in their connectome could be linked to their cognitive capacities. Finally, we link this analysis with a basic but often overlooked fact: human cognition is likely limited by our own brain structure. However, this limitation does not apply to machine learning systems, and we should be aware of the need to learn how to exploit this augmented vision of the nature. 

\end{abstract}


\section{Introduction}

In a previous work (Marco de Lucas, 2023) we have proposed that our brain processes the information through different neural networks, extracting the relevant information and storing it in a compressed form, following an architecture similar to an autoencoder, a key structure used in machine learning, and supporting the possible existence of "concept cells" (Quiroga, 2012). 
We have also discussed (Peña et al., 2024) how a simple XOR neuronal motif, built with excitatory and inhibitory neurons, could implement the analogy of a loss function in computational neural networks, solving the credit assignment problem following only a basic homeostatic criterion.
Although this previous work was applied to basic engrams, following a model based in the simple C. Elegans neurons, it can be extended to models with more complex architectures, including those with spiking neurons, as it has been done in its computational counterpart (Bidollahkhani et al., 2023) for liquid time constant (LTC) neural networks (Hasani et al., 2021).

In this new work we will analyse how the structure of the latent space of these neural networks may have an impact on the cognitive potential. First, we will show how the complexity of the data being processed must be related to the structure of the neural network that processes it, and in a first approximation to the dimension of its latent space. We will then propose how a biological system may build an index on top of this latent space, and how different concepts could be linked as parts of an engram to compose episodic memory. We will discuss the implications on the cognitive potential of different species, comparing some basic ideas about their connectomes. Finally, we will reflect on the implications of a structural limit in the capacity of the human brain, which does not apply to machine learning systems.

\section{Structure of an autoencoder and intrinsic dimension of data}

The processing of information in a neural network following the architecture of an autoencoder, and trained in unsupervised mode, is organized in different sections (see figure 1): 

-The encoder section, that handles the incoming data and processes it through a series of “hidden layers”, extracting the most relevant features of each data to be able to identify it correctly when a new instance is presented.

-The latent space, a layer where the output of the encoder section is connected, with a reduced dimension.

-The decoder section, that enables recovering an “image” of the data, from a given point in the latent space. 

The decoder section may be completely symmetrical to the encoder one, but not necessarily: the decoder may simply provide a recovery image with the detail required for the usage expected, however it must provide sufficient details to identify uniquely a new incoming data when processed by the encoder section, to define if it matches or not a previous data image.

\begin{figure}[h!]
	\centering
       \adjustbox{cfbox=black 1pt}{ 
       \includegraphics[width=0.6\textwidth]{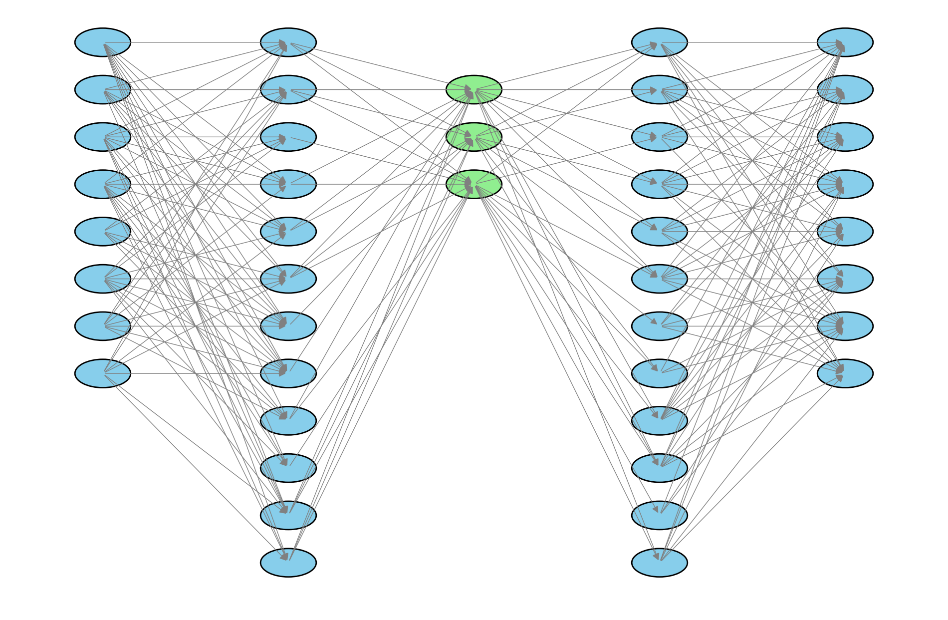} }
	\caption{Architecture of a basic neuronal network autoencoder. Latent space neurons are drawn in green.}
	\label{fig:autoencoder}
\end{figure}

The information can then be considered stored in these three layers, as weights in the connection of the neural nodes, and also including the internal state values of the neural nodes in the case of recurrent neural networks. 

Next, we introduce a concept that is as simple as nontrivial: the intrinsic dimension of data. An initial definition was given in (Bennet, 1965): “the minimum number of parameters (or degrees of freedom) necessary to capture the entire information present in the representation “, that can be stated equivalently in more precise terms as “the dimensionality of the $m$-dimensional manifold M embedded within the $d$-dimensional ambient (representation) space P where $ m \leq d$ .” (Gong et al., 2019). The renewed interest in this concept comes from “the fundamental geometric property of a data representation in a neural network as the minimal number of coordinates which are necessary to describe its points without significant information loss” (Ansuini et al., 2019).

Computational neural networks are usually overparametrized, and can also be pruned without losing performance, and that may be likely also the case for the biological neural networks, so there is not a one to one correspondence between information dimension and the dimension of the latent space of an autoencoder, but it is clear that the dimension of the latent space must be at least equal or greater than the dimension of the information it encodes. 

Notice that the dimension of the latent space itself is not a direct concept: it can be as simple as an n-dimensional vector for a basic autoencoder, but it may involve a deeper non-linear structure, even involving an internal recurrent structure. We will restrict in our considerations to the simplest case, where each data that is processed by the neural network is encoded into an n-dimensional vector in the latent space. 

For example, if we are processing the data in ImageNET (Deng et al., 2009), i.e. several million images with a typical size 224x224x3 pixels, and we have estimated that the intrinsic dimension of these data is 50, we could build an autoencoder where any image in such collection is represented by a single n-tuple of real numbers with dimension 50. 

There are many relevant points concerning spaces of very large dimension, that are both non-intuitive and related. The first one is that in such spaces, in a “n-dimensional sphere” the majority of the points are close to the “surface” of such sphere. This simple fact indicates that it is relatively “easy” to separate such points. Along the same direction, it can be proved that if the data represented in a n-dimensional space corresponding to the same category is compact, then one (Vinyals et al., 2017) or few shot (Snell et al., 2017) learning is feasible (Tyukin et al., 2021). This is a key point from a biological perspective, where learning should not require a large number of iterations, nor massive training data. And it is also important to consider that it is possible to use “quantized” n-dimensional spaces, i.e. spaces where the coordinates are not required to be real numbers, but rather discrete values.

\section{Building a latent space with biological feasibility}

The implementation of a latent space of large dimension (N) with biological neurons can adopt different forms. The simplest is to configure a set of N “independent” neurons, where each one configures a dimension axis through its dendrites (see figure 2). 

\begin{figure}[h!]
      \centering
       \adjustbox{cfbox=black 1pt}{ 
      \includegraphics[width=0.8\textwidth]{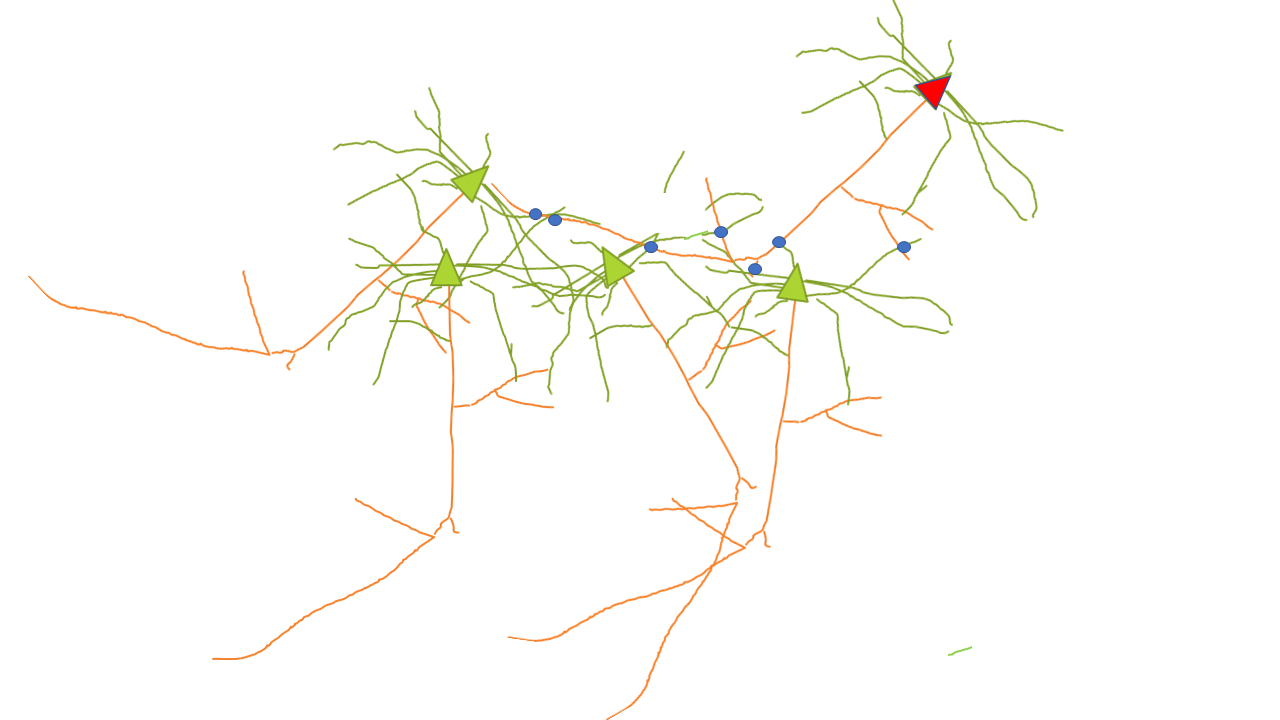}}
      \caption{ Proposed layout from a pyramidal 'concept' neuron (soma as red triangle) to four pyramidal neurons (soma in green) in the latent space. For illustrative purposes, axons are drawn in orange and dendrites in green, while potential synapses are marked as blue small circles. It should be noted that the concept neuron has the possibility of multiple axon-dendrite connections towards the neurons configuring the latent space. This simplified graph does not reflect the complexity of these very dense 3D networks, nor the role of other neuron types, in particular inhibitory ones.}
	\label{fig:LatentSpaceNeurons}
\end{figure}

The values along this axis may be considered as quantized if we count only the existence of a connection in a given dendrite. Notice that for large excitatory neurons the number of possible connections can range up to several thousands, although typically are in the order of hundredths. For example, a simple latent space of dimension 50 to encode images would require a set of 50 neurons. In this simple case, each image could be “indexed” over this latent space by a separate (“concept”) neuron, whose axon is connected to different dendrites in the 50 neurons. When this neuron is activated, these 50 neurons may be activated or not through these synapses, providing an encoded representation of the image that can be translated through the decoder into the original image. To provide a typical figure, a pyramidal cell axon from a human superior temporal gyrus with a path length of 3.74 mm, may establish 132 output synapses (Loomba et al., 2022).

This is a very basic, didactic, scheme, as there may exist multiple synapses between the index neuron and the latent space neurons, and we have not considered the complex intra-dendritic structure. The situation is even more complex if the autoencoder is based in a recurrent neural network, and if it involves both excitatory and inhibitory neurons, like in the basic example described in (Peña et al, 2024).

The question is if this hypothesis makes sense for the complex information corresponding to the nature that we, and any other animal species, observe and process. The complexity of the real world perceived is in fact constrained by two factors: the sensory system on one side, and the neuronal and motor system on the other. Guided by evolutionary principles, these systems are as complex as required to satisfy the basic question: how to survive. Even if our eyes have an incredible visual capacity, our brain does not consider all minor details of all the images seen.

Regarding images, the analysis described in (Pope et al., 2021) using the 20M images from ImageNET, classified into 20.000 categories, indicates that the intrinsic dimension of such data may be as low as 50.

Two very different studies pointed also in this direction. The study presented in (Chang and Tsao, 2017) on the code for facial identity in the primate brain, stated that a set of only 50 neurons was triggered on this task. In parallel, the Facenet paper (Schroff, Kalenichenko and Philbin, 2015) demonstrated how a computational neural network with a limited latent space dimension ( $< 50$ ), was able to provide an adequate identification of human faces. As recognition of faces is one of the most relevant tasks for social relationships, and key to guarantee survival, it is expected that these neuronal circuits are among the most evolved and likely complex in our brains.  

The potential structure of the latent space must be reflected in the connectome, i.e. the graph describing the connections (synapses) between neurons and, if possible, their type (excitatory or inhibitory). As a first example, the complete connectome of the C. Elegans is well known (see WormWiring), although the excitatory or inhibitory type of some synapses is yet uncertain (Rakowski et al., 2013). As checked in (Peña et al., 2024), this connectome includes many XOR motifs that could be linked to the existence of a recurrent neural network with a small latent space configured around only 4 neurons.

Regarding the brain in mammals, the studies published about connectomes correspond to small sections, but they provide very valuable information related to a potential latent space. The study from (Reimann et al., 2017) using rat neocortical micro circuitry, checks the dimension of cliques, i.e. complete subgraphs, of neurons within the connectome, to find a maximum value of 15 in the dimension of these substructures, that aggregate to compose larger structures. This fact points to a either a sparse arrangement of a latent space, or to a relatively low limit of its dimension.  

Two more recent studies (Loomba et al., 2022) (Hunt et al., 2023), analyse in detail the differences observed between the neuronal structures in the brain of mice/rats, macaques and humans. The main factors discussed regarding the impact of the neuronal plasticity on the different cognitive capacity, include the dimension of neuronal structures as graphs, the stability and complexity of the synapsis, and the relevance of inhibitory structures. A very recent paper (Kanari et al., 2024) puts the emphasis on the increase of dendritic complexity, but it also shows a clear difference in the network complexity between mice and humans.

\section{From concept neurons towards episodic memory}

As shown before, the existence of a latent space for a neuronal circuit processing a type of information, for example visual or auditive, opens the possibility to build a concept neuron, connecting these different latent spaces by simple Hebbian association. Also, recent advances in computational neural networks on embedding spaces (Girdhar et al., 2023) shows the path to build multimodal engrams (Artiles et al., 2023).

It has been proposed that these multimodal concept neurons may be found in the hippocampus (Quiroga, 2023), and the proposal described in this paper supports this hypothesis, hopefully clarifying previous objections (Barwich, 2019), and sharing the complexity of the learning tasks with other brain areas, connecting the corresponding latent spaces to the concept neurons in the hippocampus.

However, this is not the end of the story. Hippocampus is also proposed to manage the episodic memory (Kolibius et al., 2023), and other types of more complex mental structures, like for example different types of conceptual maps.
 
The neural network architecture proposed here, with recurrent neural networks built using biological neurons, has a direct connection with structured state space models (Gu, Goel and Ré, 2022), that have shown a remarkable plasticity in the Long Range Arena benchmark (Tay et al., 2020), including PathX, where human performance is much better than most current AI models, including transformers. The Deep Linear Recurrent architecture introduced in (Orvieto et al., 2023) stacks linear recurrent layers with nonlinear multilayer perceptron layers, matching this performance on long range paths, and offers a clear inspiration to explore a similar structure in biological neural networks, and in particular in the vision system of mammals. 

An episodic memory following a similar structure would be able to build not only stories, like it is possible with transformers, but also to follow complex reasoning paths. Additionally, as a network, it could even exploit a Boolean logical architecture, using both the XOR and the NOT (inhibitory) neuronal motifs to state implications (i.e., if P then Q, where P and Q are propositions, i.e, sets of concepts), without the need to explore all potential combinations in data to select the most likely (like it is done in current large language models). However, it will be difficult to test all these hypotheses without a deeper knowledge of the connectome, and also of the exact way that information is encoded and processed in our spiking neurons.

\section{Cognitive implications, conclusions and outlook}

This short note analyses, mainly with a didactic purpose, the relevance of the structure of the latent space conformed by a limited set of neurons in the brain, if autoencoders are the basic mechanism used to learn and store the information as engrams.  

The number and complexity of these neuronal connections would be related to the intrinsic dimension of the data analysed, i.e. to the perception of the external world, and the capacity to interact with it.
  
Following the analogy with computational neural networks, the cognitive implications are quite direct: this structure establishes a clear limit. For example, if the dimension of the visual latent space is limited to 15, it is unlikely that a animal brain will be able to differentiate the images that as humans we can identify and classify. On the other hand, if the human brain visual latent space has a typical dimension limited to, for example,  a value around 100, it will not be able to differentiate/recognize images if the intrinsic dimension of these images in nature exceeds this limit. The existence of this limit is often overlooked, but it is obvious if we compare the cognition capacities of the different animal species, and how they have evolved.
 
If the reason for this limit is linked to the dimension and complexity of the latent spaces in the neural networks, it does not apply to machine learning in computing systems, where ideally, we can build systems as complex as desired, limited only by hardware evolution.  To exploit the extraordinary capacities of these systems, handling the information in these very large n-dimensional latent spaces, it is likely that we would need to develop an effective theory, as our brain has already done to explore the nature with our senses. This would be a very challenging, but attractive, task.

\end{document}